\documentclass[11pt]{article}
\usepackage[preprint]{acl}
\usepackage{times,latexsym}
\usepackage[T1]{fontenc}
\usepackage[utf8]{inputenc}
\usepackage{microtype,inconsolata}
\usepackage{amsmath,amssymb,amsthm}
\usepackage{booktabs,array,tabularx,graphicx,xcolor,placeins}
\newcommand{\NA}{\textsc{Abstain}}
\newcommand{\LF}{\textsc{LocalFeedback}}
\newtheorem{proposition}{Proposition}
\hypersetup{hidelinks,pdftitle={Curbing Error Propagation in Reasoning Agents via Scoped Arithmetic Feedback},pdfauthor={Guangxin Su; Siyuan Su}}
\title{Curbing Error Propagation in Reasoning Agents\\via Scoped Arithmetic Feedback}
\author{
  Guangxin Su \\
  UNSW \\
  \texttt{guangxinsu6@gmail.com}
  \And
  Siyuan Su\thanks{Conducted most of the experiments.} \\
  Yanshan University \\
  \texttt{19563983824@163.com}
}

\begin{document}
\maketitle
\begin{abstract}
An exact arithmetic refutation does not establish that a reasoning agent has asserted an error: the expression also occurs in quotations and rejected assumptions. We studied scoped arithmetic feedback, an interface that binds exact calculations to source spans and separately evaluates their discourse role. Its messages authorize local reconsideration without certifying an answer or deleting a reasoning branch. Paired revision experiments covered three generators from two model families on GSM8K and a Qwen2.5-7B extension on MATH-500. Across the four cohorts, scoped routing reduced rendered feedback records by 76.9--80.0\% and produced zero observed corruptions; raw routing corrupted one to three initially correct answers per cohort. Two-sided 95\% Clopper--Pearson upper endpoints for scoped corruption were 2.9--10.9\%. Raw feedback also repaired more answers in every cohort. Final accuracy differed by zero or one question, with exact McNemar \mbox{$p=1.00$} throughout. Separately, a 3,400-trace ProcessBench routing analysis showed that uncertainty or binding failures accounted for 82 of 98 suppressed refutations. Conditional alignment increased while first-error recall decreased. These findings exposed a consistent repair--retention tradeoff within the evaluated cohorts and separated arithmetic validity, feedback volume, and answer-level outcomes without treating zero observed corruption as a general guarantee.
\end{abstract}

\section{Introduction}
A calculator evaluates an expression; a reasoning agent interprets its role in a solution. These judgments diverge in ordinary mathematical discourse. A proof by contradiction contains a false equality that the argument explicitly rejects. A quotation reproduces a claim without endorsing it. A corrected derivation retains an earlier mistake in its history. In each case, exact arithmetic alone leaves the appropriateness of corrective feedback undetermined.

Tool-assisted revision combines several judgments: what the response asserts, whether extraction preserves that assertion, whether the formal expression holds, and how the agent uses the result. Process reward models estimate step quality \citep{lightman2024verify}; executable arithmetic supplies deterministic evidence about explicitly parsed inputs. Reliability requires an interface that preserves this distinction when a tool result becomes a language intervention.

We study \emph{scoped arithmetic feedback} for agents that generate and revise mathematical solutions. The interface binds a calculation to its question, response, and source span; attaches a separate assertion/boundary label; and emits a local feedback record or an abstention. Its authority remains local: a refuted equality supplies a reason to reconsider that calculation, without certifying the full answer or deleting a reasoning branch.

This work makes three contributions. First, we implement an agent-feedback interface that separates exact arithmetic replay, source binding, discourse interpretation, and local critique. Second, we analyze fixed-support routing on ProcessBench and attribute every suppressed refutation to its discourse label, uncertainty status, or binding failure. Third, we examine repair and corruption separately in a 200-question GSM8K pilot and three 100-question extensions spanning two additional generators and MATH-500. Scoped feedback produced zero observed corruptions in each cohort, while raw feedback repaired more initially wrong answers. Final correctness differed by at most one question. This decomposition exposed a repair--retention tradeoff across the tested conditions.

Together, trace-level routing analysis and paired answer revision connect the interface mechanism to its coverage and repair--retention tradeoffs.

\begin{figure*}[t]
\centering
\includegraphics[width=\textwidth]{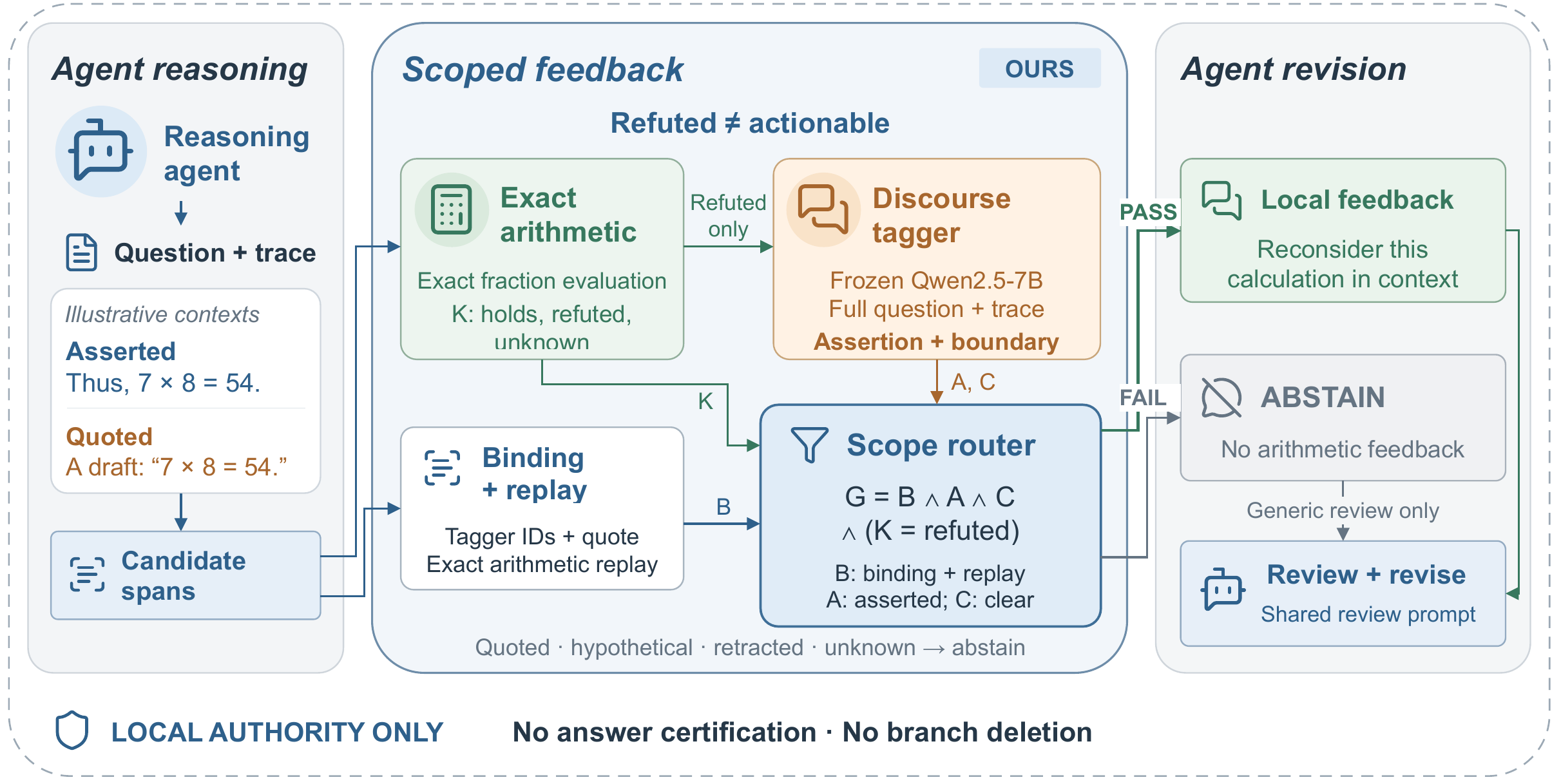}
\caption{Exact refutation is not sufficient for actionable feedback. Illustrative contexts show the same false equality as an assertion and a quotation. The scoped route combines exact arithmetic, fallible discourse labels, and replay bound to source text and tagger records; only refuted candidates enter discourse tagging. The gate authorizes a local calculation critique or abstains from arithmetic feedback. Both paths retain generic review. The dashed boundary excludes answer certification and branch deletion.}
\label{fig:scoped-feedback-architecture}
\end{figure*}

\section{Related Work}
\paragraph{Tool-augmented reasoning and revision.}
ReAct interleaves reasoning with actions \citep{yao2023react}, and PAL delegates computation to executable programs \citep{gao2023pal}. Self-Refine uses generated feedback for iterative revision \citep{madaan2023selfrefine}; CRITIC incorporates external tools into critique \citep{gou2024critic}. Intrinsic self-correction alone remains unreliable on reasoning tasks \citep{huang2024selfcorrect}. These approaches motivate external evidence, but executing a calculation and deciding whether its result warrants a critique are distinct judgments. A false equality inside a quotation or rejected assumption does not identify an error endorsed by the current solution. Our interface addresses this intervention decision and evaluates both repairs and corruptions, which final accuracy combines.

\paragraph{Process supervision and step-level verification.}
Process reward models learn step-quality scores from human supervision \citep{lightman2024verify}; self-consistency selects answers across sampled reasoning paths \citep{wang2023selfconsistency}. ProcessBench evaluates localization of the earliest reasoning error \citep{zheng2025processbench}. These methods address semantic reasoning beyond a restricted arithmetic language. Exact arithmetic instead supplies a replayable verdict under explicit parsing semantics, with no independent conclusion about assertion status or the final answer. We therefore distinguish local arithmetic validity, first-error localization, and revision outcomes. A route that retains fewer well-aligned alerts improves conditional alignment while reducing error recall, as our controlled comparison demonstrates.

\paragraph{Discourse scope and attribution.}
FactBank models event factuality as expressed in discourse, distinguishing actual, nonactual, and uncertain interpretations \citep{sauri2009factbank}. PARC 3.0 annotates attribution through source, cue, and content spans \citep{pareti2016parc}. These distinctions motivate treating the occurrence of an expression separately from a speaker's commitment to it. Our assertion labels adapt this perspective to generated mathematical reasoning, including quotations, hypotheses, and retractions. Source binding checks the identity and location of a tagger's evidence; it does not establish that the tag is correct. The local authority contract preserves this boundary between deterministic arithmetic and fallible discourse interpretation.

\paragraph{Selective prediction and feedback abstention.}
Selective classification trades prediction coverage against error risk by rejecting uncertain inputs \citep{geifman2017selective}. Our system applies abstention to feedback emission: unsupported arithmetic, invalid bindings, and nonasserted or uncertain scope withhold a critique while retaining generic review. This decision differs from refusing to answer the task. We measure its coverage and repair--retention effects directly. The frozen discourse tagger has no calibrated risk guarantee, so fewer emitted messages and zero observed corruptions remain empirical outcomes rather than certified semantic reliability.

\section{Scoped Arithmetic Feedback}
\label{sec:method}
Figure~\ref{fig:scoped-feedback-architecture} highlights the distinction between a false equality and actionable feedback: discourse scope and source-bound replay determine which local critiques reach the agent.
\subsection{Evidence and language interpretation}
Let $q$ be a question and $y$ an agent response. An extractor returns candidate spans \mbox{$s=(a,b,y[a:b])$}. The arithmetic checker maps a candidate to
\begin{equation}
K(q,y,s)\in\{\mathsf{holds},\mathsf{refuted},\mathsf{unknown}\}.
\end{equation}
A record stores the question and response digests, offsets, literal text, parsed expression, exact values, checker identity, and outcome. Replay checks these bindings and recomputes the arithmetic. Unknown covers unsupported syntax, undefined operations, and resource limits; it is not a judgment that the reasoning is wrong.

The reference checker parses bounded integer, decimal, and rational expressions, including explicit arithmetic operators and equality chains. It evaluates them with exact fractions. It extracts standalone equation lines and delimited mathematical spans. Variables, units, approximations, and implicit multiplication lie outside this language. A decimal is interpreted as an exact rational literal, not as an unspecified approximation. Lexical context vetoes in the frozen parser are fallible filters, not discourse annotations. Parser bounds and assurance checks appear in Appendix~\ref{app:artifacts}.

A separate annotation $z(s)$ records assertion status and boundary status. Assertion status distinguishes \emph{asserted}, \emph{quoted}, \emph{hypothetical}, \emph{retracted}, and \emph{unknown}; the boundary is \emph{clear} or \emph{ambiguous}. Both question and response identities bind the annotation. These labels express an annotator's or model's interpretation. Their semantic validity is separate from arithmetic replay.

\begin{table}[t]
\centering\small
\begin{tabularx}{\columnwidth}{@{}X l@{}}
\toprule
Context of a refuted expression & Route \\
\midrule
Asserted; clear boundary; current source & Local feedback \\
Quoted, hypothetical, or retracted & Abstain \\
Ambiguous boundary or stale annotation & Abstain \\
Undefined or unsupported arithmetic & Abstain \\
\bottomrule
\end{tabularx}
\caption{Routing decisions. Assertion tags remain fallible. Neither route certifies an answer or deletes a reasoning branch.}
\label{tab:route}
\end{table}

\subsection{Routing and revision}
Define $B$ as successful evidence replay and annotation binding, $A$ as an asserted label, and $C$ as a clear boundary. The gate and feedback policy are
\begin{equation}
\begin{aligned}
G&=B\land A\land C\land(K=\mathsf{refuted}),\\
R&=
\begin{cases}
\LF,&G,\\
\NA,&\text{otherwise}.
\end{cases}
\end{aligned}
\label{eq:route}
\end{equation}
The implementation additionally abstains when a span begins with a spaced list marker that is indistinguishable from a unary operator. It does not silently strip that marker and change the claim. A stale question or response annotation also triggers abstention. A checker-identity mismatch raises an error; the calling agent retains the response and records the failure.

\subsection{Automatic scope and agent feedback}
The runtime requests discourse labels from a frozen Qwen2.5 7B instruction model using the full question, response, and refuted candidates, without an answer key. The schema requires a candidate identifier, assertion and boundary tags, and a literal evidence quotation. Runtime code locates a unique quotation in the source and derives its offsets. A duplicated or missing identifier, malformed record, truncated generation, or unlocatable evidence leads to abstention. A valid schema binds an interpretation to text; it does not prove that interpretation.

Only refutations enter semantic annotation; other arithmetic outcomes return no feedback. This avoids charging scope inference to unsupported claims. The tagger remains fixed across generators. Each feedback record quotes the complete original span and states the exact values of the first unequal adjacent pair in an equality chain. It asks the agent to reconsider the calculation in context and supplies no target answer. Raw and scoped feedback use the same renderer; they differ only in which records pass the route.

The renderer takes at most eight complete records in source order within a 1,800-byte UTF-8 budget. It never truncates an equation to fit; omitted records are counted. Empty feedback gives the exact generic review prompt. The companion revision runtime implements generic, raw-arithmetic, and scoped-feedback arms with a shared initial response and review seed. The ProcessBench analysis evaluates feedback routing without invoking revision. The implementation records model calls, feedback omissions, abstentions, tokens, and stop reasons. Implementation checks and their scope are documented in Appendix~\ref{app:artifacts}.

\subsection{What the contract guarantees}
\begin{proposition}[Local arithmetic scope]
Assume the parser/evaluator implements its specified exact semantics and replay binds the record to the current inputs. Every emitted feedback record refers to a false equality under those parsed semantics. This property alone entails neither that the passage asserts the equality nor that the answer is incorrect.
\end{proposition}
\noindent\emph{Argument.} The routing condition requires a replayed refutation, so the parsed values differ. Assertion status and task grounding do not occur in the arithmetic judgment. A false equality inside a quotation, or one already retracted later in the response, supplies a counterexample to the stronger conclusion.

This is a permission contract, not a new arithmetic theorem. Even an asserted mistake such as \mbox{$7\times8=54$} admits the revision \mbox{$7\times8=56$}. Local refutation therefore provides no exclusion proof over all possible continuations. The interface exposes neither answer-acceptance nor branch-deletion authority. Its trusted base comprises the parser, exact-number runtime, binding checks, and enforcement of the route; hashes detect changes, not truth about the problem.

\section{Evaluation}
\label{sec:design}
We evaluated scope routing on ProcessBench and paired answer revision on GSM8K and MATH-500. Traces and questions defined the reporting units; claims described behavior within those units.

\subsection{ProcessBench fixed-support replay}
The comparison covered 3,400 traces from 2,842 distinct question strings. We retained the original step-local extraction and all arithmetic certificates. The frozen Qwen2.5 7B tagger received the complete question and trace, with checked mappings from evidence quotations to the original spans. All 122 refutations entered scope processing; inputs without valid bound labels retained abstention. Traces without a raw refutation retained their no-alert outcome. The intervention changed routing while preserving the parser's language and calculations.

Published labels identified the first reasoning error and all-correct traces, without assessing every equation. We classified the first emitted alert as before, at, or after the annotated first error. First-alert alignment used alerted traces as its denominator; first-error recall used all 2,221 error-labeled traces. Alerts on 1,179 all-correct traces formed a separate outcome. Multiple claims within one trace were not independent observations.

Error-attribution and routing analysis linked each original refutation to its predicted label, binding status, and emitted feedback. We reconstructed paired first-alert transitions for all 69 originally alerted traces, checking input identities and saved certificates before aggregation. The analysis introduced no new model responses or human labels. Inference costs included failed requests and retained input and output tokens separately.

\subsection{Paired 200-question revision pilot}
\label{sec:pilot-protocol}
The pilot used 200 GSM8K test questions \citep{cobbe2021verifiers}, selected by sorting SHA-256 digests of question strings. Qwen2.5-7B-Instruct used Q4\_K\_M quantization. Initial generations used temperature 0.7, top-$p$ 0.95, and seed 17. Each question shared one initial response across generic review, raw arithmetic feedback, and scoped feedback. Each arm received one review with a shared review seed, a common renderer, and a 1,024-token output cap. Review order was hash-seeded. The temperature-zero scope tagger received no answer key. The pilot comprised 200 initial responses and 600 reviews; reference answers entered scoring after generation.

The shared initial responses contained 124 correct and 76 wrong answers. Repair used the 76 initially wrong answers as its denominator; corruption used the 124 initially correct answers. We compared raw and scoped final correctness with an exact McNemar test and a paired question bootstrap. For this binary paired outcome, we computed the bootstrap distribution directly from the $2\times2$ counts, removing Monte Carlo variation. A two-sided 95\% Clopper--Pearson interval accompanied the zero-corruption count. Appendix~\ref{app:pilot200} details the paired transition counts and interval calculations.

\subsection{Generator and dataset extensions}
\label{sec:extensions-protocol}
Two additional GSM8K cohorts evaluated Qwen2.5-14B-Instruct and Llama-3.1-8B-Instruct on 100 questions each. These generators tested a larger Qwen model and a different model family. A fourth cohort evaluated Qwen2.5-7B-Instruct on the first 100 MATH-500 questions in the evaluation order. This benchmark is the 500-problem MATH test subset released with \citet{lightman2024verify}, using the Hugging Face H4 distribution \citep{math500}. Each cohort used three paired review arms with a shared initial response and review seed within a question. The cross-cohort tables report raw and scoped outcomes; generic review provides the primary pilot's reference comparison.

We retained each cohort's initial-correct and initial-wrong denominators and reported corruption intervals separately. The extension analyses were descriptive: they did not pool models, assume that question sets were disjoint, or test architectural invariance. All CP upper endpoints refer to marginal, two-sided 95\% Clopper--Pearson intervals. Paired corruption tests used exact McNemar comparisons of the shared initially correct responses. Appendix~\ref{app:extension-statistics} gives the paired inference supported by the reported margins.

\begin{table*}[t]
\centering\small
\renewcommand{\arraystretch}{1.16}
\begin{tabular*}{\textwidth}{@{\extracolsep{\fill}}llrrrrrr@{}}
\toprule
 & & & & \multicolumn{3}{c}{First alert vs. first error} & \\
\cmidrule(lr){5-7}
Split (traces) & Route & \shortstack{Emitted\\claims} & \shortstack{Alerted\\traces} & At & Before & After & \shortstack{On correct\\traces} \\
\midrule
GSM8K (400) & Raw & 9 & 2 & 1 & 0 & 1 & 0/193 \\
 & Scoped & 1 & 1 & 1 & 0 & 0 & 0/193 \\
\addlinespace[2pt]
MATH (1,000) & Raw & 47 & 24 & 11 & 2 & 11 & 0/406 \\
 & Scoped & 10 & 10 & 6 & 0 & 4 & 0/406 \\
\addlinespace[2pt]
OlympiadBench (1,000) & Raw & 24 & 14 & 5 & 0 & 8 & 1/339 \\
 & Scoped & 3 & 3 & 1 & 0 & 2 & 0/339 \\
\addlinespace[2pt]
Omni-MATH (1,000) & Raw & 42 & 29 & 9 & 2 & 14 & 4/241 \\
 & Scoped & 10 & 10 & 3 & 0 & 6 & 1/241 \\
\addlinespace[2pt]
\midrule
All (3,400) & Raw & 122 & 69 & 26 & 4 & 34 & 5/1179 \\
 & Scoped & 24 & 24 & 11 & 0 & 12 & 1/1179 \\
\bottomrule
\end{tabular*}
\caption{Controlled routing comparison on ProcessBench with identical extracted candidates and arithmetic evidence. Emitted claims counts local refutations; At, Before, and After locate the first alert relative to the annotated first error. The final column gives alerted all-correct traces over all-correct traces. These are trace-localization counts, not equation-level precision; routing preserves the arithmetic evaluations.}
\label{tab:acl-pb}
\end{table*}

\section{Results}
\label{sec:diagnostics}
\subsection{Scope routing: what the filter removed}
The frozen parser evaluated 2,217 equalities among 91,940 candidates: 2,095 held, 122 were refuted, and 89,723 returned unknown. Supported claims occurred in 814 traces. These counts established the exact arithmetic coverage of the diagnostic; they did not estimate the prevalence of reasoning errors.

Scoped routing emitted feedback for 24 of the 122 refutations. Table~\ref{tab:acl10-routes} decomposes the remaining 98. Eleven received a quoted label and five a retracted label; none received a hypothetical label. Fifty-two had source-bound unknown labels, nine became unknown after evidence-quotation validation failed, and 21 lacked bound labels. Consequently, uncertainty and binding failures accounted for 82/98 suppressed claims. Explicit quotation or retraction predictions accounted for 16/98. The dominant observed mechanism was selective abstention, with a smaller contribution from typed discourse exclusions.

\begin{table}[t]
\centering\small
\begin{tabularx}{\columnwidth}{@{}Xr@{}}
\toprule
Recorded route & Claims \\
\midrule
Asserted, clear; feedback emitted & 24 \\
Quoted; abstained & 11 \\
Retracted; abstained & 5 \\
Hypothetical; abstained & 0 \\
Source-bound unknown; abstained & 52 \\
Invalid evidence; downgraded to unknown & 9 \\
Unbound labels; abstained & 21 \\
\midrule
Total raw refutations & 122 \\
\bottomrule
\end{tabularx}
\caption{Error-attribution analysis of the frozen scope route. Assertion types are model predictions, not human validity labels. A source binding verifies the quotation, not its discourse interpretation.}
\label{tab:acl10-routes}
\end{table}

The distinction mattered for interpreting the apparent reduction in false alarms. Of the four all-correct traces whose alerts disappeared, three were suppressed through unknown labels and one through an unbound record. None of these four reductions was attributable to a quoted or retracted label. Moreover, source binding did not establish that a tag was semantically correct: in a MATH factorization example, the response presented \mbox{$168=2^3\times7$} as a prime factorization, but the tagger labeled it quoted. The equality was exactly refuted and its quotation was correctly bound; the interpretation still suppressed a directly stated calculation. Appendix~\ref{app:qualitative} contrasts this error with a retained arithmetic critique and a correctly suppressed contradiction. These cases explain routing decisions without estimating their population frequencies.

\subsection{ProcessBench: selectivity and localization}
The scoped route reduced alerted traces from 69 to 24. First-alert alignment increased from 26/69 (37.7\%) to 11/24 (45.8\%), while first-error recall decreased from 26/2,221 (1.17\%) to 11/2,221 (0.50\%). Alerts on all-correct traces decreased from 5/1,179 to 1/1,179. Table~\ref{tab:acl-pb} retains the dataset-specific counts. The relative reduction of four out of five correct-trace alerts described a small observed cell; it did not establish elimination of false feedback or preservation of error recall.

Reconstructing the paired transitions clarified how these aggregates arose. All 24 retained first alerts stayed in the same relation to the published error label: 11 coincided with it, 12 followed it, and one belonged to an all-correct trace. The route suppressed 15 previously aligned alerts, 22 later alerts, four earlier alerts, and four correct-trace alerts. It did not turn an earlier or later first alert into an aligned one. Therefore, the higher conditional alignment ratio reflected selection among existing alerts rather than improved first-error localization within a trace.

The fixed arithmetic support also bounded achievable first-error recall. Only 27 traces contained a raw refutation at the annotated first-error step: one GSM8K, 12 MATH, five OlympiadBench, and nine Omni-MATH traces. The raw first alert already aligned on 26; only one trace from the \mbox{MATH} split supplied an additional aligned candidate after an earlier alert. Any route restricted to selecting these candidates therefore had a first-error recall ceiling of 27/2,221 (1.22\%). This was a support bound, independent of the accuracy of a discourse classifier.

\subsection{Observed inference cost}
All 69 originally alerted traces entered scope processing. The runtime issued 68 tagger calls; one trace exceeded the conservative full-context guard and retained an abstention. Sixty traces produced complete valid label records, and eight failed the exact candidate-ID check. The tagger consumed 126,029 input and 7,537 output tokens. These measured costs included scope processing on traces that ultimately produced no feedback.

The measured output of this expenditure was a smaller alert set with changed localization coverage. ProcessBench contained no agent revisions, so these costs did not yield a cost per successful repair or a reduction in revision loops. The paired revision cohorts below evaluated answer transitions separately.

\subsection{Dynamic pilot: repair and retention}
\label{sec:dynamic-pilot}
Table~\ref{tab:pilot} reports the paired revision outcomes. Raw feedback reached 126/200 final correctness (63.0\%), with 5/76 repairs (6.6\%) and 3/124 corruptions (2.4\%). Scoped feedback reached 125/200 (62.5\%), with 1/76 repairs (1.3\%) and 0/124 corruptions. Generic review also reached 125/200, with two repairs and one corruption. Equal final scores therefore concealed different balances between fixing errors and preserving correct answers.

\begin{table}[t]
\centering\small
\begin{tabular}{@{}lrrr@{}}
\toprule
Arm & Correct/200 & Repair/76 & Corrupt/124 \\
\midrule
Generic & 125 & 2 & 1 \\
Raw & 126 & 5 & 3 \\
Scoped & 125 & 1 & 0 \\
\bottomrule
\end{tabular}
\caption{Paired GSM8K pilot. Each arm revised the same 200 initial responses; 124 were correct and 76 were wrong. Cells are counts with denominators in the headings.}
\label{tab:pilot}
\end{table}

\begin{table}[t]
\centering\small
\begin{tabular}{@{}lrrr@{}}
\toprule
 & Scoped correct & Scoped wrong & Total \\
\midrule
Raw correct & 122 & 4 & 126 \\
Raw wrong & 3 & 71 & 74 \\
\midrule
Total & 125 & 75 & 200 \\
\bottomrule
\end{tabular}
\caption{Paired final correctness. The seven discordant pairs give exact McNemar \mbox{$p=1.00$}. The raw-minus-scoped difference is $+0.5$ pp (paired bootstrap 95\% interval $[-2.0,+3.0]$ pp).}
\label{tab:mcnemar}
\end{table}

The raw-minus-scoped accuracy difference was $+0.5$ percentage points (paired bootstrap 95\% interval $[-2.0,+3.0]$). Table~\ref{tab:mcnemar} preserves all 200 paired outcomes. Although final accuracy did not differ significantly (exact McNemar \mbox{$p=1.00$}), the observed retention outcomes differed: scoped feedback preserved all 124 initially correct answers, whereas raw feedback corrupted three (2.4\%). Zero observed corruption had a two-sided 95\% Clopper--Pearson interval of $[0.0,2.9]\%$. The paired corruption comparison itself gave \mbox{$p=0.25$}; these counts describe the observed tradeoff without establishing a population-level reduction in corruption.

\paragraph{Paired answer transitions.}
The seven discordant outcomes comprised three initially correct answers retained only by scoped review and four initially wrong answers repaired only by raw review. Both arms retained the other 121 initially correct answers and repaired one initially wrong answer. The paired table therefore separates retention from repair even when their net effects nearly cancel.

\paragraph{Feedback volume and exposure.}
Scoped feedback reduced rendered records from 45 to 9 and rendered UTF-8 bytes from 12,480 to 2,496, both by 80\%. Initially wrong responses receiving feedback decreased from 31/76 to 7/76, a 77.4\% relative reduction. The lower repair count accompanied this lower exposure. These measurements establish a reduction in feedback volume, not an 80\% reduction in semantically irrelevant content.

\subsection{Cross-generator and cross-dataset outcomes}
\label{sec:extension-results}
\begin{table*}[t]
\centering\small
\setlength{\tabcolsep}{4pt}
\begin{tabular*}{\textwidth}{@{\extracolsep{\fill}}llrrrrrrr@{}}
\toprule
Dataset & Generator & $N$ & Initial & Final & Repair & Corrupt & Records & Reduction \\
 & & & correct & R/S & R/S & R/S & R/S & records \\
\midrule
GSM8K & Qwen2.5-7B & 200 & 124 & 126/125 & 5/1 & 3/0 & 45/9 & 80.0\% \\
GSM8K & Qwen2.5-14B & 100 & 78 & 80/79 & 3/1 & 1/0 & 26/6 & 76.9\% \\
GSM8K & Llama-3.1-8B & 100 & 58 & 59/59 & 3/1 & 2/0 & 36/8 & 77.8\% \\
MATH-500 & Qwen2.5-7B & 100 & 32 & 34/33 & 3/1 & 1/0 & 44/10 & 77.3\% \\
\bottomrule
\end{tabular*}
\caption{Paired revision across generators and datasets. R/S denotes raw/scoped counts, not a ratio. Final correctness uses $N$; corruption uses initially correct answers; repair uses initially wrong answers ($N-$initial correct). Repair counts follow from the initial, final, and corruption marginals. Record reductions measure rendered feedback volume. Each cohort has exact final-accuracy McNemar \mbox{$p=1.00$}; cohorts are not pooled.}
\label{tab:extensions}
\end{table*}

Table~\ref{tab:extensions} reports all four cohorts with their own denominators. On GSM8K, the 14B generator started with 78/100 correct answers: raw review corrupted one and repaired three, while scoped review corrupted none and repaired one. Final correctness was 80/100 versus 79/100. Llama-3.1-8B started with 58/100 correct answers: raw review corrupted two and repaired three, while scoped review corrupted none and repaired one. Both arms ended at 59/100. Thus, the extension to a second model family reproduced the separation between answer retention and repair despite equal final accuracy.

The zero-corruption cells for the three GSM8K generators had CP upper endpoints of 2.9\%, 4.6\%, and 6.2\%. The direction of the observed comparison was consistent across generators, but the wider extension intervals reflected smaller initial-correct denominators. The observations did not establish a generator-independent corruption rate. Rendered record counts fell from 26 to 6 for Qwen2.5-14B (76.9\%) and from 36 to 8 for Llama-3.1-8B (77.8\%).

On MATH-500, Qwen2.5-7B started with 32/100 correct answers. Raw review repaired 3/68 initially wrong answers and corrupted 1/32 initially correct answers, ending at 34/100. Scoped review repaired 1/68 and corrupted 0/32, ending at 33/100. Records fell from 44 to 10 (77.3\%). The scoped corruption upper endpoint was 10.9\%, reflecting the 32-answer retention denominator. This cohort supplied a cross-dataset observation, with substantially greater uncertainty than the primary GSM8K cohort. Its final-accuracy comparison also gave \mbox{$p=1.00$}; it remained separate from the GSM8K results.

\section{Discussion}
\paragraph{Final accuracy hides the repair--retention tradeoff.}
For initial accuracy $a_0$, repair rate $r$ among initially wrong answers, and corruption rate $c$ among initially correct answers, accuracy after revision satisfies
\begin{equation}
 a_1=a_0(1-c)+(1-a_0)r.
\end{equation}
Raw review repaired more initially wrong answers in every cohort: five versus one in the primary pilot and three versus one in each extension. Scoped review instead preserved every initially correct answer in the observed samples. These opposing transitions left zero- or one-question differences in final correctness. Nonsignificant accuracy tests did not establish equivalence, non-inferiority, or unchanged repair ability. The observed benefit was answer retention with less rendered feedback, accompanied by lower repair counts.

\paragraph{Breadth and uncertainty have separate roles.}
The extensions covered three generators from two model families and two mathematical benchmarks. Their zero scoped-corruption counts were directionally consistent, while their upper confidence endpoints ranged from 2.9\% to 10.9\%. The model-specific denominators determined these intervals. The extensions broadened the observed conditions; they did not reject a small-sample explanation for zero events or establish stability under arbitrary generator or distribution changes. We retained separate cohorts and did not treat shared questions or models as independent replications of one pooled rate.

\paragraph{Selectivity is not semantic precision.}
On ProcessBench, uncertainty and binding failures accounted for 82 of 98 suppressed claims, including all four removed correct-trace alerts. Explicit quotation and retraction labels accounted for the other 16. Every retained first alert stayed in its original localization category, and fifteen aligned alerts disappeared. Higher conditional alignment therefore reflected selection with a coverage cost. The factorization case in Appendix~\ref{app:qualitative} also showed that source binding did not prevent an incorrect scope label from suppressing an asserted error. ProcessBench explained routing behavior; it supplied no dynamic corruption estimate. Likewise, the 76.9--80.0\% reduction in rendered records measured message volume, not a labeled semantic signal-to-noise ratio.

\paragraph{Cost includes the gate as well as the message.}
The ProcessBench tagger consumed 68 calls and 133,566 tokens, including processing that ended in abstention. The primary GSM8K pilot reduced rendered feedback bytes by 80\%; extension tables reported record counts. Neither quantity measured total model tokens, latency, or energy. The primary pilot gave every arm one review, with no observed savings in revision loops. Scope inference and generator input belong in the same end-to-end cost account; shorter feedback alone establishes no efficiency advantage.

\paragraph{Local authority remained intact.}
Arithmetic replay established that parsed values differed, and binding established which text supplied them. Neither operation certified assertion status, quantity grounding, or whole-answer correctness. The semantic gate selected a bounded critique, and the agent retained control of its answer. This separation made the repair, retention, and coverage tradeoffs auditable without extending a local verdict into answer acceptance or branch deletion.

\FloatBarrier
\section{Conclusion}
Scoped arithmetic feedback separated exact computation from the decision to intervene on an agent's reasoning. Across three generators on GSM8K and a MATH-500 extension, scoped routing reduced rendered records by 76.9--80.0\% and preserved all initially correct answers in each observed cohort. Raw routing repaired more answers, leaving final-accuracy differences of zero or one question. The ProcessBench routing analysis attributed most suppression to selective abstention and exposed its localization cost. The central result was a repair--retention tradeoff across the evaluated conditions, with explicit uncertainty and local feedback authority.

\label{main-text-end}
\section*{Limitations}
The checker evaluated parsed rational arithmetic. Extraction fidelity, quantity grounding, discourse interpretation, and whole-answer correctness remained separate judgments. The parser, exact-number runtime, and binding implementation formed the trusted base; finite tests and mutation checks did not constitute whole-program verification.

The automatic route used one frozen quantized scope tagger. Schema validation and unique evidence quotations established source identity, not semantic validity. The conservative context guard and binding failures contributed to the reported coverage loss and remained in the denominators. ProcessBench supplied first-error labels, not equation-level human validity annotations.

The dynamic evaluation comprised a 200-question primary GSM8K cohort, two 100-question generator extensions on GSM8K, and one 100-question MATH-500 extension. The three generators represented two model families, not three independent architectures. These small cohorts did not estimate retraining or repeated-run variation. MATH-500 was evaluated on its first 100 questions in the evaluation order. Each corruption interval was marginal, with no simultaneous cross-cohort coverage guarantee. Zero observed corruption retained nonzero upper endpoints, and the paired comparisons established neither a significant corruption reduction nor accuracy equivalence. Aggregate counts supported outcome decompositions but did not identify question-level feedback mechanisms. English mathematical reasoning defined the domain; multilingual reasoning, open-domain factual claims, and interactive tool environments lay outside it.

\section*{Ethical Considerations}
The system emitted local critiques and abstentions rather than whole-answer safety or correctness guarantees. The analyses used public benchmark traces and questions and treated their contents as data. Benchmark reference answers served as scoring targets rather than feedback content. Release materials retained source-dataset and model attribution and licensing information and excluded credentials and private user data.

\bibliography{references,references_acl}

@inproceedings{wang2023selfconsistency,
  title={Self-Consistency Improves Chain of Thought Reasoning in Language Models},
  author={Wang, Xuezhi and Wei, Jason and Schuurmans, Dale and Le, Quoc V. and Chi, Ed H. and Narang, Sharan and Chowdhery, Aakanksha and Zhou, Denny},
  booktitle={International Conference on Learning Representations},
  year={2023}
}

@inproceedings{lightman2024verify,
  title={Let's Verify Step by Step},
  author={Lightman, Hunter and Kosaraju, Vineet and Burda, Yuri and Edwards, Harrison and Baker, Bowen and Lee, Teddy and Leike, Jan and Schulman, John and Sutskever, Ilya and Cobbe, Karl},
  booktitle={International Conference on Learning Representations},
  year={2024},
  url={https://proceedings.iclr.cc/paper_files/paper/2024/hash/aca97732e30bcf1303bc22ac3924fd16-Abstract-Conference.html}
}

@inproceedings{yao2023react,
 title={{ReAct}: Synergizing Reasoning and Acting in Language Models},
 author={Yao, Shunyu and Zhao, Jeffrey and Yu, Dian and Du, Nan and Shafran, Izhak and Narasimhan, Karthik and Cao, Yuan},
 booktitle={International Conference on Learning Representations}, year={2023},
 url={https://openreview.net/forum?id=WE_vluYUL-X}}

@inproceedings{gao2023pal,
 title={{PAL}: Program-aided Language Models},
 author={Gao, Luyu and Madaan, Aman and Zhou, Shuyan and Alon, Uri and Liu, Pengfei and Yang, Yiming and Callan, Jamie and Neubig, Graham},
 booktitle={Proceedings of the 40th International Conference on Machine Learning}, year={2023}, pages={10764--10799},
 url={https://proceedings.mlr.press/v202/gao23f.html}}

@article{cobbe2021verifiers,
 title={Training Verifiers to Solve Math Word Problems},
 author={Cobbe, Karl and Kosaraju, Vineet and Bavarian, Mohammad and Chen, Mark and Jun, Heewoo and Kaiser, Lukasz and Plappert, Matthias and Tworek, Jerry and Hilton, Jacob and Nakano, Reiichiro and Hesse, Christopher and Schulman, John},
 journal={arXiv preprint arXiv:2110.14168}, year={2021}, url={https://arxiv.org/abs/2110.14168}}

@inproceedings{zheng2025processbench,
    title = "{P}rocess{B}ench: Identifying Process Errors in Mathematical Reasoning",
    author = "Zheng, Chujie  and
      Zhang, Zhenru  and
      Zhang, Beichen  and
      Lin, Runji  and
      Lu, Keming  and
      Yu, Bowen  and
      Liu, Dayiheng  and
      Zhou, Jingren  and
      Lin, Junyang",
    editor = "Che, Wanxiang  and
      Nabende, Joyce  and
      Shutova, Ekaterina  and
      Pilehvar, Mohammad Taher",
    booktitle = "Proceedings of the 63rd Annual Meeting of the {A}ssociation for {C}omputational {L}inguistics (Volume 1: Long Papers)",
    month = jul,
    year = "2025",
    address = "Vienna, Austria",
    publisher = "Association for Computational Linguistics",
    url = "https://aclanthology.org/2025.acl-long.50/",
    doi = "10.18653/v1/2025.acl-long.50",
    pages = "1009--1024",
    ISBN = "979-8-89176-251-0"
}

@inproceedings{madaan2023selfrefine,
 author = {Madaan, Aman and Tandon, Niket and Gupta, Prakhar and Hallinan, Skyler and Gao, Luyu and Wiegreffe, Sarah and Alon, Uri and Dziri, Nouha and Prabhumoye, Shrimai and Yang, Yiming and Gupta, Shashank and Majumder, Bodhisattwa Prasad and Hermann, Katherine and Welleck, Sean and Yazdanbakhsh, Amir and Clark, Peter},
 booktitle = {Advances in Neural Information Processing Systems},
 doi = {10.52202/075280-2019},
 editor = {A. Oh and T. Naumann and A. Globerson and K. Saenko and M. Hardt and S. Levine},
 pages = {46534--46594},
 publisher = {Curran Associates, Inc.},
 title = {{Self-Refine}: Iterative Refinement with Self-Feedback},
 url = {https://proceedings.neurips.cc/paper_files/paper/2023/file/91edff07232fb1b55a505a9e9f6c0ff3-Paper-Conference.pdf},
 volume = {36},
 year = {2023}
}

@inproceedings{gou2024critic,
 title={{CRITIC}: Large Language Models Can Self-Correct with Tool-Interactive Critiquing},
 author={Gou, Zhibin and Shao, Zhihong and Gong, Yeyun and Shen, Yelong and Yang, Yujiu and Duan, Nan and Chen, Weizhu},
 booktitle={International Conference on Learning Representations}, year={2024},
 url={https://arxiv.org/abs/2305.11738}}

@inproceedings{huang2024selfcorrect,
 title={Large Language Models Cannot Self-Correct Reasoning Yet},
 author={Huang, Jie and Chen, Xinyun and Mishra, Swaroop and Zheng, Huaixiu Steven and Yu, Adams Wei and Song, Xinying and Zhou, Denny},
 booktitle={International Conference on Learning Representations}, year={2024},
 url={https://arxiv.org/abs/2310.01798}}

@misc{math500,
 title={{MATH-500}},
 author={{Hugging Face H4}},
 year={2024},
 howpublished={\url{https://huggingface.co/datasets/HuggingFaceH4/MATH-500}},
 note={Dataset distribution of the 500-problem MATH test subset released with \emph{Let's Verify Step by Step}},
 url={https://huggingface.co/datasets/HuggingFaceH4/MATH-500}
}

@article{sauri2009factbank,
 title={{FactBank}: A Corpus Annotated with Event Factuality},
 author={Saur{\'i}, Roser and Pustejovsky, James},
 journal={Language Resources and Evaluation},
 volume={43}, number={3}, pages={227--268}, year={2009},
 doi={10.1007/s10579-009-9089-9},
 url={https://doi.org/10.1007/s10579-009-9089-9}
}

@inproceedings{pareti2016parc,
    title = "{PARC} 3.0: A Corpus of Attribution Relations",
    author = "Pareti, Silvia",
    editor = "Calzolari, Nicoletta  and
      Choukri, Khalid  and
      Declerck, Thierry  and
      Goggi, Sara  and
      Grobelnik, Marko  and
      Maegaard, Bente  and
      Mariani, Joseph  and
      Mazo, Helene  and
      Moreno, Asuncion  and
      Odijk, Jan  and
      Piperidis, Stelios",
    booktitle = "Proceedings of the Tenth International Conference on Language Resources and Evaluation ({LREC}'16)",
    month = may,
    year = "2016",
    address = "Portoro{\v{z}}, Slovenia",
    publisher = "European Language Resources Association (ELRA)",
    url = "https://aclanthology.org/L16-1619/",
    pages = "3914--3920"
}

@inproceedings{geifman2017selective,
 author = {Geifman, Yonatan and El-Yaniv, Ran},
 booktitle = {Advances in Neural Information Processing Systems},
 editor = {I. Guyon and U. Von Luxburg and S. Bengio and H. Wallach and R. Fergus and S. Vishwanathan and R. Garnett},
 publisher = {Curran Associates, Inc.},
 title = {Selective Classification for Deep Neural Networks},
 url = {https://proceedings.neurips.cc/paper_files/paper/2017/file/4a8423d5e91fda00bb7e46540e2b0cf1-Paper.pdf},
 volume = {30},
 year = {2017}
}
\clearpage
\appendix
\section{Recorded Experimental Configuration}
\label{app:protocol}
\paragraph{Automatic scope run.}
ProcessBench scope inference used \texttt{qwen2.5:7b}, Q4\_K\_M quantization, and temperature zero. The manifest bound the model digest, runtime, source, inputs, and saved requests and responses. Full questions and traces entered the tagger. The context guard produced one abstention without an API call; 68 calls included eight candidate-ID validation failures.

\paragraph{Binding and rendering.}
Each label supplied a candidate ID, assertion/boundary tags, and a unique source quotation. Missing or duplicated IDs rejected the response-level label set. Invalid evidence downgraded the label to unknown. Runtime checks bound question, response, and span identities before routing.

The renderer retained at most eight complete records in source order within 1,800 UTF-8 bytes, recording whole-record omissions. Each record quoted its source and the exact values of the first unequal adjacent pair, followed by: ``This checks only the quoted calculation. Reconsider its role in your solution.'' Empty feedback added no message. Runtime token counts remained distinct from this byte limit.

\section{Evaluation Pipeline and Checker Validation}
\label{app:artifacts}
\paragraph{Offline analysis pipeline.}
The analysis uses four deterministic stages: candidate extraction and exact arithmetic evaluation, quotation-uniqueness replay validation, route-category attribution, and first-error support coverage analysis. Saved discourse predictions supply the input to replay; inference itself remains a separate learned stage. Each stage preserves the question, response, and span identities, ensuring that comparisons use the same source text and arithmetic evidence.

Paired revision separates initial generation, feedback routing and rendering, and final-answer scoring. Generation receives the question without a reference answer; evaluation compares completed responses with the reference. Within a question, all review arms share the initial response and review seed. Pilot and extension statistics are computed from the paired outcome counts described in Appendices~\ref{app:pilot200} and~\ref{app:extension-statistics}.

\paragraph{Checker validation.}
The checker limits each candidate to 512 characters, 128 tokens, 64-digit numeric literals, parse depth 24, 2,048-bit intermediate values, and integer exponents from $-12$ to $12$. A dedicated parser excludes unrestricted program evaluation. Differential checks compare 3,721 defined expressions with an independent integer-pair reference and evaluate 7,442 true/false certificates and 56 boundary cases. Thirteen of fourteen nonequivalent semantic mutants are detected; the surviving mutant changes a depth limit outside the tested corpus. Undefined expressions are excluded from the equivalence count. These finite checks characterize implementation behavior rather than proving the checker correct for every input.

\paragraph{Error localization and support.}
ProcessBench scoring associates completed route predictions with the published first-error labels. The loader has access to these labels, but the tagger prompt omits them. We analyze 91,940 candidate records and count a trace as supporting first-error localization when at least one raw refutation occurs at its annotated first-error step. Filtering existing candidates cannot introduce an alert at an unsupported step. The resulting 27-trace ceiling is a coverage bound, independent of discourse-classifier accuracy. The routing analysis uses fixed generated traces; it does not measure subsequent answer revision.

\section{Paired Pilot Configuration and Statistics}
\label{app:pilot200}
Section~\ref{sec:pilot-protocol} gives the dataset selection, model, sampling, and pairing settings. Review generations used temperature 0.7 and top-$p$ 0.95. The scope tagger received no reference answer.

The renderer retained at most eight complete records in source order within 1,800 UTF-8 bytes; it never truncated an equation. An empty route supplied the generic review prompt. Generic review rendered no arithmetic records. Raw review rendered 45 records and 12,480 bytes; scoped review rendered 9 records and 2,496 bytes. The counts of initially wrong responses exposed to feedback were 0, 31, and 7 for generic, raw, and scoped review, respectively.

The primary GSM8K pilot's answer extractor selected a final explicit answer field, then a sentence with an answer cue among the final three sentences if that field was absent. A bounded numeric grammar mapped the answer to a rational. Incompatible multiple numbers were marked ambiguous and scored as incorrect. Reference answers entered scoring after generation.

\paragraph{Paired transition counts.}
The paired $2\times2$ counts were $(122,4,3,71)$: both correct, raw-only correct, scoped-only correct, and both incorrect. They total 200, with correct marginals of 126 and 125. The transition identities were \mbox{$124-3+5=126$} (raw), \mbox{$124-0+1=125$} (scoped), and \mbox{$124-1+2=125$} (generic). These counts give the outcome decompositions in Section~\ref{sec:dynamic-pilot}; attribution of individual revisions to feedback messages is a separate analysis.

\paragraph{Paired accuracy inference.}
For question $i$, let $d_i$ be raw correctness minus scoped correctness. Its empirical probabilities at \mbox{$d_i=-1,\,0,\,+1$} were, respectively,
\[
 \left(\frac{3}{200},\;\frac{193}{200},\;\frac{4}{200}\right).
\]
A 200-fold convolution gives the paired nonparametric bootstrap distribution of the summed difference. Its 2.5\% and 97.5\% quantiles were $-4$ and $6$ questions, or \mbox{$[-2.0,+3.0]$} percentage points. Direct evaluation eliminates Monte Carlo error without guaranteeing exact finite-sample coverage. The four-versus-three discordances give two-sided exact McNemar \mbox{$p=1.00$}.

\paragraph{Corruption uncertainty.}
With zero corruptions among 124 initially correct answers, the two-sided 95\% Clopper--Pearson interval had upper endpoint
\[
 1-0.025^{(1/124)}\approx0.02931.
\]
The paired comparison had three raw-only and zero scoped-only corruptions, yielding exact McNemar \mbox{$p=0.25$}. No observed corruption does not imply zero population risk.

\section{Extension Statistics and Reporting Units}
\label{app:extension-statistics}
\begin{table}[t]
\centering\small
\begin{tabular}{@{}lrrrr@{}}
\toprule
Cohort & Initial & Raw & Scoped & Paired \\
 & correct & upper (\%) & upper (\%) & $p$ \\
\midrule
Q7/GSM & 124 & 6.9 & 2.9 & 0.25 \\
Q14/GSM & 78 & 6.9 & 4.6 & 1.00 \\
L8/GSM & 58 & 11.9 & 6.2 & 0.50 \\
Q7/MATH & 32 & 16.2 & 10.9 & 1.00 \\
\bottomrule
\end{tabular}
\caption{Corruption uncertainty by cohort. Upper endpoints belong to marginal, two-sided 95\% Clopper--Pearson intervals. Paired $p$ values are unadjusted, two-sided exact McNemar tests for corruption, not final accuracy. Q7/Q14/L8 abbreviate the generators in Table~\ref{tab:extensions}.}
\label{tab:corruption-intervals}
\end{table}

For each cohort and arm, let $I$, $F$, and $C$ count initially correct answers, finally correct answers, and corruptions. The repair count is
\begin{equation}
 R=F-I+C.
\end{equation}
Applying this identity gives three raw repairs and one scoped repair in each extension, with initial-wrong denominators of 22, 42, and 68. Record reductions and corruption intervals use the corresponding cohort-level counts; no pooled rate combines different generators or datasets.

\paragraph{Paired tests.}
For final correctness, let $b$ denote raw-only correct and $c$ scoped-only correct. Their difference equals the final-correctness marginal difference. For every compatible table with \mbox{$\lvert b-c\rvert\leq1$}, the two-sided exact McNemar test returns \mbox{$p=1.00$}, because the observed discordance is central under a symmetric binomial null. Thus, the reported extension $p$ values follow from the given margins without reconstructing individual pairings. Their total discordances and confidence intervals for accuracy differences remain undetermined by those margins.

For corruption, zero scoped events fix the discordant counts: $(3,0)$, $(1,0)$, $(2,0)$, and $(1,0)$ across the four cohorts. The corresponding two-sided exact McNemar values are 0.25, 1.00, 0.50, and 1.00. Fisher's independent-group calculation does not retain the matched-response structure and is not the primary test. No cohort or model was pooled into a common McNemar statistic.

\paragraph{Intervals and measurement scope.}
Table~\ref{tab:corruption-intervals} uses the same two-sided 95\% Clopper--Pearson convention throughout. Its upper endpoints are not one-sided 95\% bounds or a simultaneous guarantee over four cohorts. Record counts describe rendered messages; extension byte totals are not inferred from the primary pilot's 80\% byte reduction. The reused 200-question row is a reference cohort, not an additional replication. Cross-generator summaries retain their cohort-specific denominators and make no assumption that the GSM8K question sets are disjoint.

\section{Qualitative Routing Analysis}
\label{app:qualitative}
\begin{figure*}[t]
\centering
\setlength{\fboxsep}{7pt}
\fbox{\begin{minipage}[t]{0.94\textwidth}
\small
\textbf{A. An asserted error receives feedback (GSM8K, example 36).}\\
\textbf{Context.} The solution sums the number of beach visitors across a week.\\
\textbf{Trace excerpt.} ``Now, we sum the number of people''\quad $50+70+140+240=400$.\\
\textbf{Exact check.} The left side is $500$; the right side is $400$.\\
\textbf{Raw / scoped.} Both emit the local critique. The tagger predicts \emph{asserted} with a clear boundary, and the critique coincides with the annotated first-error step. No final-answer certificate follows from this local verdict.
\end{minipage}}\par\medskip
\fbox{\begin{minipage}[t]{0.94\textwidth}
\small
\textbf{B. An explicitly rejected contradiction is suppressed (OlympiadBench, example 101).}\\
\textbf{Context.} A candidate case in a system of equations produces $-1-1-1=1$, followed by $-3=1$.\\
\textbf{Trace excerpt.} ``This is a contradiction, so there is no solution in this case.''\\
\textbf{Exact check.} Both equalities compare $-3$ with $1$.\\
\textbf{Raw / scoped.} Raw emits two local refutations; scoped emits none after \emph{retracted} labels. The passage already rejects these equalities. Suppression concerns their local discourse role and does not validate the preceding derivation or the complete solution.
\end{minipage}}\par\medskip
\fbox{\begin{minipage}[t]{0.94\textwidth}
\small
\textbf{C. A mistaken quotation label suppresses an asserted error (MATH, example 19).}\\
\textbf{Context.} The solution uses prime factorization to compute the greatest common factor of $180$ and $168$.\\
\textbf{Trace excerpt.} ``Second,''\quad $168=2^3\times7$.\\
\textbf{Exact check.} The two sides evaluate to $168$ and $56$.\\
\textbf{Raw / scoped.} Raw emits a critique at the annotated first-error step. Scoped abstains because the tagger predicts \emph{quoted}, despite the response directly presenting a factorization. The evidence quotation is correctly bound to the source; the semantic interpretation is wrong. Binding therefore does not certify assertion status.
\end{minipage}}
\caption{Qualitative comparison of local feedback routes on three observed traces. Short source excerpts are reproduced; surrounding task descriptions are condensed. Cases were selected to illustrate distinct routing behaviors, not to estimate their frequency or explain the dynamic corruption counts.}
\label{fig:routing-cases}
\end{figure*}

Figure~\ref{fig:routing-cases} illustrates retained, suppressed, and misclassified arithmetic feedback in three observed ProcessBench traces.

\end{document}